\documentclass[letterpaper, 10 pt, conference]{ieeeconf}  

\usepackage{cite}
\usepackage{amsmath,amssymb,amsfonts}
\usepackage{algorithmic}
\usepackage{graphicx}
\usepackage{textcomp}
\usepackage{xcolor}
\usepackage{nccmath}
\usepackage{float}
\usepackage{multirow}
\usepackage{subfig}

\IEEEoverridecommandlockouts                              

\title{\LARGE \bf
Optimal Design of Continuum Robots with Reachability Constraints
}

\author{Hyunmin Cheong$^{1}$, Mehran Ebrahimi$^{1}$,
        and Timothy Duggan$^{2}$%
\thanks{This work is partly supported by the Office of Naval Research contract \#N68335-17-C-0045, ARPA-E contract \#DE-AR0001241, and NASA contract \#80NSSC19C0637.}%
\thanks{$^{1}$Hyunmin Cheong and Mehran Ebrahimi are Research Scientists at Autodesk Research, Autodesk Inc. 
{\tt \footnotesize hyunmin.cheong@autodesk.com, mehran.ebrahimi@autodesk.com}}%
\thanks{$^{2}$Timothy Duggan is the Soft Robotics Research Co-Lead at Otherlab Inc. {\tt \footnotesize tduggan@otherlab.com}}
}

\begin{document}

\maketitle

\thispagestyle{empty}
\pagestyle{empty}

\begin{abstract}
While multi-joint continuum robots are highly dexterous and flexible, designing an optimal robot can be challenging due to its kinematics involving curvatures. Hence, the current work presents a computational method developed to find optimal designs of continuum robots given reachability constraints. First, we leverage both forward and inverse kinematic computations to perform reachability analysis in an efficient yet accurate manner. While implementing inverse kinematics, we also integrate torque minimization at joints such that robot configurations with the minimum actuator torque required to reach a given workspace could be found. Lastly, we apply an estimation of distribution algorithm (EDA) to find optimal robot dimensions while considering reachability, where the objective function could be the total length of the robot or the actuator torque required to operate the robot. Through three application problems, we show that the EDA is superior to a genetic algorithm (GA) in finding better solutions within a given number of iterations, as the objective values of the best solutions found by the EDA are 4-15\% lower than those found by the GA.


\end{abstract}


\section{Introduction}

\begin{figure}[bt!]
\centering
\includegraphics[width=1.72in]{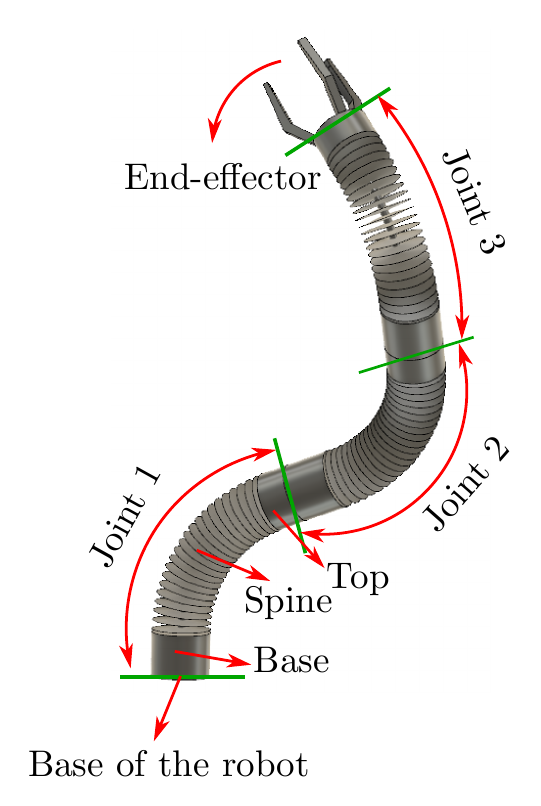}
\caption{A three-joint spatial continuum robot. Each joint consists of a base, top and spine. The spine can bend continuously along its length about two orthogonal axes.}
\label{fig:robot3d}
\end{figure}

Continuum robots, also called soft robots, are composed of joints that bend continuously along their lengths (Fig. \ref{fig:robot3d}). The design of such manipulators are inspired by animal appendages such as elephant trunks or octopus tentacles. They are highly dexterous and flexible, thus ideal for working in cluttered environments. Furthermore, compared to traditional rigid manipulators, continuum robots can better adapt to and interact with their surroundings. However, due to their flexibility, analyzing their reachability requires a more complex kinematic model than traditional manipulators and it is difficult to conceptualize the effect of design changes on the workspace of these robots. Even with 3D computer-aided design tools, designing a continuum robot that can reach the entire target workspace can take weeks of an engineering team's time, and the chance of finding an optimal design (e.g., with the minimal total length) is low.

To address this challenge, we present a method for optimizing the design of multi-joint continuum robots while satisfying reachability constraints given a desired workspace. The method involves two main parts: 1) kinematic computations to evaluate the reachability of potential designs and 2) an optimization algorithm to find the optimal robot design. 

For the first part, we combine both forward and inverse kinematics to perform a reachability analysis in an efficient manner. Since the former can be executed considerably faster than the latter, a large set of randomly sampled robot configurations with forward kinematics is used to quickly estimate the robot's reachability. Then, for the points not deemed to be reachable, inverse kinematics is employed to check whether those points can actually be reached or not, further improving the accuracy of the reachability analysis. In addition, we integrate torque minimization as part of the inverse kinematics computation such that given a target point, the robot configuration with the minimal actuator torque to reach that point can be identified. 

The optimization problem considering reachability is challenging because computing the sensitivity of the reachability function with respect to the design variables involved is not possible. Also, the feasible region is likely non-contiguous since different combinations of robot dimensions could satisfy the reachability constraint. To solve such a problem, derivative-free or black-box optimization algorithms can be used, e.g., evolutionary algorithms. For the current work, we use an estimation of distribution algorithm (EDA), which finds optimal solutions by estimating and sampling a probability model of promising designs. EDA is chosen because it has been shown to outperform a genetic algorithm, the most widely used evolutionary algorithm, in a number of prior studies \cite{ pelikan1999bivariate, shakya2007optimization, shakya2012markovianity, martins2018performance, cheong2019configuration}. Also, the problem knowledge in the form of probabilistic models learned as part of the optimization could be conveyed to the user, which is not an inherent feature of most evolutionary algorithms.

The rest of the paper is organized as follows. First, we present related work on continuum robots and EDAs. Next, a problem formulation and the method developed to solve the problem are presented. We then provide the results of the experiments conducted to validate our optimization method, followed by a summary and conclusions.

\section{Related Work}

\subsection{Kinematics of Continuum Robots}
Unlike traditional robot manipulators, continuum robots are made of sequentially actuated deformable links (joints) exhibiting some levels of compliance depending on their design and the applications destined for them. Therefore, the conventional parameters (e.g., joint lengths and joint angles) used for describing the kinematics of rigid-link robots are not applicable to continuum robots. A thorough review of different techniques for modeling the kinematics of such robots can be found in \cite{trivedi2008soft, transeth2009survey, webster2010design}. A common practice adopted in analyzing continuum robots is to assume that each of their joints deforms as a constant-curvature arc (e.g. \cite{hannan2003kinematics, jones2006kinematics, neppalli2007design, kapadia2013self, garriga2019kinematics, allen2020closed}). This turns the robot's configuration space from an infinite to a finite dimensional space and facilitates fast closed-form computations of the robot kinematics. Following this presumption, several kinmeatic models have been proposed.  In \cite{hannan2003kinematics}, the well-known Denavit-Hartenberg procedure is modified to determine the kinematics of planar robots resembling an elephant's trunk. Another approach is to model each continuum joint as a combination of prismatic and revolute joints \cite{jones2006kinematics}. In \cite{kapadia2013self}, the motion of planar robots is characterized by splitting it into pure bending and extension. The robot's kinematics can also be configured by incorporating quaternions to define the rotation of the robot's joints \cite{garriga2019kinematics}. Despite the inherent differences among the various kinematic models, it is demonstrated that they often lead to identical results in many different scenarios \cite{webster2010design}. In the current work, the robot's three-dimensional kinematics is determined using its joints' length, radius of curvature and angle, thus bearing some similarities to what is used in \cite{webster2010design}.

\subsection{Optimal Design of Continuum Robots}

Optimal design of continuum robots has been actively researched in the past few years. Early on, much of the work was on optimizing concentric tube robots for biomedical applications. These robots are similar to the robots considered in the current work as they also require constant-curvature kinematics and can involve multiple joints. For example, generalized pattern search is used to optimize robots with the minimal curvature and length while being able to reach the workspace \cite{bedell2011design, anor2011algorithms}. In another work, the Nelder-Mead algorithm is implemented to find optimal designs while considering volume-based workspaces \cite{burgner2013computational}. Also, a particle swarm algorithm is employed to perform design optimization for dual-arm concentric tube robots \cite{chikhaoui2018toward}.

Recently, evolutionary algorithms have been widely applied for optimal design of continuum robots. For instance, Hiller et al. uses a genetic algorithm to optimally distribute soft materials in a compliant structure \cite{hiller2011automatic}. Runge et al. \cite{runge2017design} incorporates a genetic algorithm to optimize a single-joint soft robot while considering its mechanics. Cheney et al. \cite{cheney2015evolving} uses a variant of the NEAT algorithm (NeuroEvolution of Augmenting Topologies) to find the optimal volumetric structure and material choice for a contiguous soft robot. Finally, most similar to our work, Bodily et al. \cite{bodily2017multi}
applies a genetic algorithm to optimize a multi-joint continuum robot considering reachability, dexterity, and manipulability. 

\subsection{Estimation of Distribution Algorithms}

EDAs are population-based, derivative-free optimization methods that use probability distributions estimated from a population of candidate solutions to sample new solutions at each iteration of optimization \cite{larranaga2002eda, hauschild2011eda}. They have been shown to be more effective at finding optimal solutions with a fewer number of function evaluations than genetic algorithms for several benchmark problems \cite{ pelikan1999bivariate, shakya2007optimization, shakya2012markovianity, martins2018performance}, hence motivating the application for the current work.

EDAs have been applied to solve various engineering design problems \cite{gao2018estimation}, including vehicle suspensions \cite{yuen2013design, cheong2019configuration} and multi-speed gearboxes \cite{simionescu2006teeth, piacentini2020multi}. In robotics, EDAs have been used to solve problems such as inverse displacement \cite{gao2017univariate}, gait generation \cite{hu2008gait, jiang2017motion}, and cable-driven parallel robot design \cite{hernandez2017design}. However, no prior work has explored applying an EDA for optimal design of multi-joint continuum robots.

For the current work, we have considered various continuous EDAs \cite{larranaga2002eda} because the design variables involved in optimization, e.g., the dimensions of joints in a continuum robot, are continuous. Specifically, we have applied a univariate normal distribution algorithm inspired by \cite{larranaga2000optimization}.

\section{Problem Formulation}

The design of a multi-joint continuum robot is defined as the following optimization problem. 
\begin{equation}
\begin{aligned}
& \underset{\mathcal{D}}{\text{min}}
& & f(\mathcal{D}) \\
& \text{s.t.}
& & d_{i, lb} \leq x_i \leq d_{i, ub} & \forall x_i \in \mathcal{D} \\
& & & \mathbf{\Theta}(\mathcal{D},\mathcal{S}) \geq \alpha
\label{eq:opps}
\end{aligned}
\end{equation}
Here, $\mathcal{D}$ represents a set of design variables $x_i$, such as the dimensions of a robot, and $\mathcal{S}$ denotes a set of state variables that define the configuration of a robot, such as each joint's radius of curvature and angle. For the objective function $f(\mathcal{D})$, the current work considers two different types -- the total length of the robot and the minimum actuator torque required to reach a given workspace. Each design variable $x_i$ has lower and upper bounds, $d_{i, lb}$ and $d_{i, ub}$, respectively, indicating the limits on robot dimensions. Finally, $\mathbf{\Theta}(\mathcal{D},\mathcal{S})$ computes the percentage of the workspace that can be reached by a robot, and this value must be greater than the threshold $\alpha$.

In order to compute $\mathbf{\Theta}(\mathcal{D},\mathcal{S})$, a given workspace is discretized into a set of target points in the 3D space. Then, its value can be calculated by
\begin{equation}
\begin{aligned}
\mathbf{\Theta}(\mathcal{D},\mathcal{S}) = \frac{N_\text{reached}}{N_\text{target}}
\end{aligned}
\end{equation}
where $N_\text{target}$ represents the number of target points in the workspace and $N_\text{reached}$ is the number of such points that can be reached by a robot. $N_\text{reached}$ is determined based on
\begin{equation}
\begin{aligned}
\|p_j - \vec{r}_e(\mathcal{D},\mathcal{S}_{j})\| \leq \epsilon \text{   for } j=1,...,m
\end{aligned}
\end{equation}
in which $p_j$ is each of the $m$ target points considered and $\vec{r}_e(\mathcal{D},\mathcal{S}_{j})$ computes the corresponding position of the robot's end-effector as shown later. $\epsilon$ is the tolerance allowed.


\section{Methods}

First presented is the formulation of kinematics used to perform a reachability analysis, followed by the optimization method using an estimation of distribution algorithm. 

\subsection{Reachability Analysis}

As shown in \cite{bodily2017multi}, to estimate the reachability of a robot, one could randomly sample a large number of robot configurations and use forward kinematics to compute the end-effector position for each configuration. Then, the set of end-effector positions is compared against the set of target points in the workspace to estimate the reachability. Note that computing forward kinematics is quite fast and therefore one could sample a large number of configurations within a given time budget. With a large enough sample size, a reasonable estimate of the reachability could be obtained.

On the other hand, one could employ inverse kinematics to assess the reachability of a robot. For instance, for each target point in the workspace, one could solve for the robot configuration that would result in the end-effector position coinciding with the target point. The percentage of the target points for which solutions are found would then indicate the reachability. While this approach can provide a more accurate assessment of the reachability, computing inverse kinematics for all target points can take a significant amount of time. 

The current work therefore uses a hybrid approach. First, random sampling of robot configurations with forward kinematics is performed. Then, for target points that have not been reached during this first step, inverse kinematics is used to check whether those points can actually be reached or not. In this section, we present the forward and inverse kinematic models employed, while the exact implementation of the hybrid approach is described in the Experiments section.

We assume that a robot consists of several non-extensible mutually tangent constant-curvature arcs that can spatially deform. Each joint is composed of a base, top and spine as depicted in Fig. \ref{fig:robot3d}. The base and top sections are straight and rigid links, whereas the spine can bend about two orthogonal axes, thus giving each joint two degrees of freedom (DOF). Depending on the joint design, one may need to model the spine as a spatial beam including its longitudinal, bending and torsional deformations altogether \cite{ebrahimi2020low}. 

\begin{figure*}[hbt!]
\centering
\subfloat[]{\includegraphics[width=0.25\textwidth]{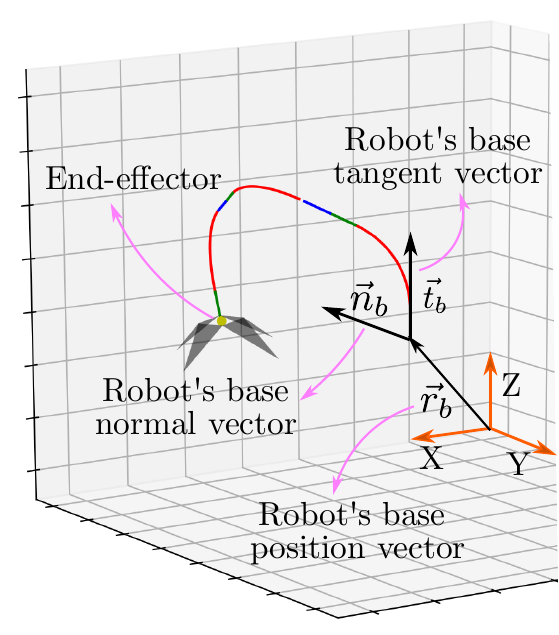}%
\label{fig:robot_2D_a}}
\subfloat[]{\includegraphics[width=0.25\textwidth]{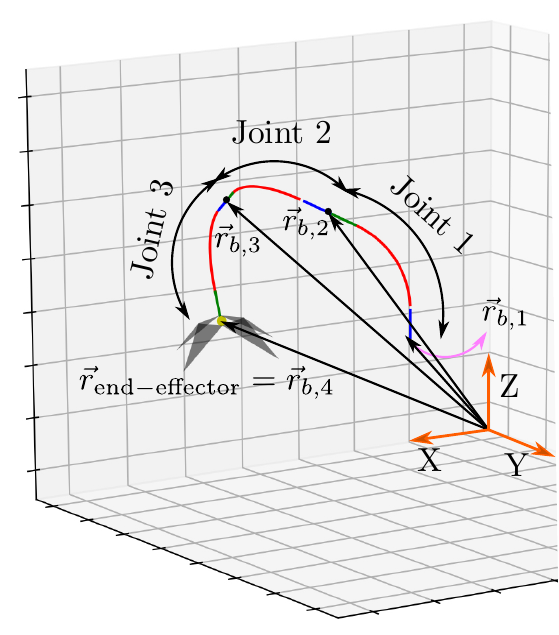}%
\label{fig:robot_2D_b}}
\subfloat[]{\includegraphics[width=0.25\textwidth]{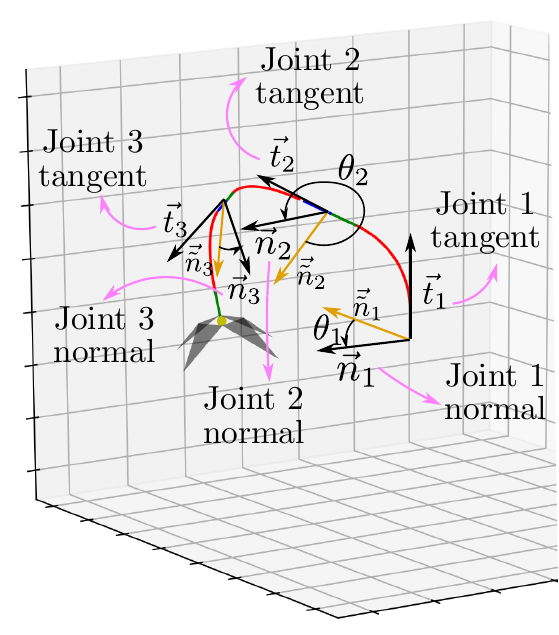}%
\label{fig:robot_2D_c}}
\subfloat[]{\includegraphics[width=0.25\textwidth]{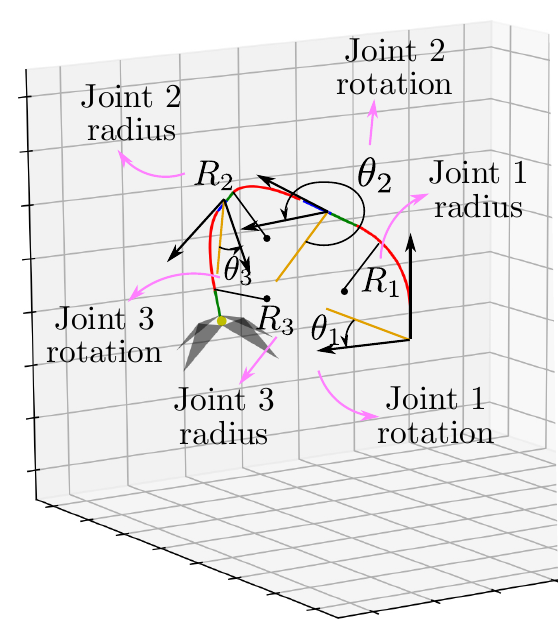}%
\label{fig:robot_2D_d}}
\caption{Simplistic representation of a three-joint continuum robot with the proposed parametrization. (a) The tangent vector at the base defines the direction towards which the first joint is pointing at, and the normal vector determines the bending plane of that joint. (b) The tip of each joint coincides with the bottom of the subsequent joint. (c) $\theta$ of each joint is defined as the rotation of $\vec{\tilde{n}}$ about that joint's tangent $\vec{t}$ leading to $\vec{n}$ determining the bending plane of that joint. (d) Given each joint's radius $R$ and $\theta$, the robot can be fully configured in space.}
\label{fig:robot_2d}
\end{figure*}

Given the robot's base position $\vec{r}_b$, tangent $\vec{t}_b$ and normal $\vec{n}_b$ (Fig. \ref{fig:robot_2D_a} and Fig. \ref{fig:robot_2D_b}), the robot kinematics can be fully configured using its joints' radius of curvature $R \: (R > 0)$ and rotation $\theta \: ( 0 \leq \theta \leq 2 \pi)$. The latter specifies the rotation of a joint with respect to its respective base normal, so it determines the bending plane of each joint. These parameters, also called state variables, are illustrated in Fig. \ref{fig:robot_2d}. In order to assess the reachability of a robot, one needs to solve the forward and inverse kinematic equations laid out as follows.

\subsubsection{Forward Kinematics}
In a forward kinematic (FK) problem, the goal is to find the spatial position and orientation at different points on the robot, particularly the end-effector, given a set of state variables (i.e. joint radii and rotations). Employing the parametrization introduced earlier and utilizing the Rodrigues' rotation formula \cite{shabana2020dynamics}, the vectors $\vec{t}_i$ and $\vec{\Tilde{n}}_i$ depicted in Fig. \ref{fig:robot_2D_c} for Joint $i$ ($i \geq 2$) read
\begin{equation}\label{eq:rntilde}
\begin{aligned}
    \vec{t}_i &= \vec{t}_{i-1} \cos \left(\phi_{i-1} \right) + \vec{n}_{i-1} \sin \left(\phi_{i-1} \right) \\
    \vec{\tilde{n}}_i &= \vec{n}_{i-1} \cos \left(\phi_{i-1}\right) - \vec{t}_{i-1} \sin \left(\phi_{i-1}\right) \\
\end{aligned}
\end{equation}
\noindent in which $\phi_{i-1} = l_{s,i-1} / R_{i-1}$; and $l_{b,i}$, $l_{s,i}$ and $l_{t,i}$ represent base, spine and top lengths of Joint $i$, respectively. Therefore, referring to Fig. \ref{fig:robot_2D_c} and using Eq. \ref{eq:rntilde} one can express $\vec{n}_i$ by
\begin{equation}\label{eq:n}
\begin{aligned}
    \vec{n}_i &= \vec{\tilde{n}}_i \cos(\theta_i) + \left(\vec{t}_i \times \vec{\tilde{n}}_i \right) \sin (\theta_i) \\
    &= \left( \vec{n}_{i-1} \cos (\phi_{i-1}) - \vec{t}_{i-1} \sin(\phi_{i-1}) \right) \cos (\theta_i) \\
    &+ \left( \vec{t}_{i-1} \times \vec{n}_{i-1} \right) \sin (\theta_i)
\end{aligned}
\end{equation}
Defining $\vec{b}_i := \vec{t}_{i} \times \vec{n}_{i}$, the three vectors $\vec{t}_{i}$, $\vec{n}_{i}$ and $\vec{b}_i$ form a Cartesian coordinate frame at the base of each Joint $i$. Through employing Eq. \ref{eq:rntilde} and Eq. \ref{eq:n}, vector $\vec{b}_{i}$ is stated as
\begin{equation}
\begin{aligned}
    \vec{b}_{i} = \vec{t}_{i} \times \vec{n}_{i} &= \left( \vec{t}_{i-1} \sin(\phi_{i-1}) - \vec{n}_{i-1} \cos(\phi_{i-1}) \right) \sin(\theta_i) \\
    &+ \vec{b}_{i-1} \cos(\theta_i)
\end{aligned}
\end{equation}
Also, for Joint 1 connected to the ground we have
\begin{equation}
    \vec{t}_1 = \vec{t}_b, \: \vec{\tilde{n}}_1 = \vec{n}_b, \:  \vec{n}_1 = \vec{\tilde{n}}_1 \cos(\theta_1) + \left(\vec{t}_1 \times \vec{\tilde{n}}_1 \right) \sin (\theta_1)
\end{equation}

The position of the bottom node of each joint $\vec{r}_{b,i} (i \geq 2)$, considering Fig. \ref{fig:robot_2d}, Eq. \ref{eq:rntilde} and Eq. \ref{eq:n} can be formulated as
\begin{equation}\label{eq:rb}
\begin{aligned}
    \vec{r}_{b,i} &= \vec{r}_{b,i-1} + l_{b,i-1} \vec{t}_{i-1} + R_{i-1} \left(\vec{n}_{i-1} - \vec{\tilde{n}}_{i} \right) + l_{t,i-1} \vec{t}_{i} \\
    &= \vec{r}_{b,i-1} \\
    &+ \left(l_{b,i-1} + R_{i-1} \sin(\phi_{i-1}) + l_{t,i-1} \cos(\phi_{i-1}) \right) \vec{t}_{i-1} \\ 
    &+ \left(R_{i-1} - R_{i-1} \cos(\phi_{i-1}) + l_{t,i-1} \sin(\phi_{i-1}) \right) \vec{n}_{i-1}
\end{aligned}
\end{equation}
\noindent with $\vec{r}_{b,1} = \vec{r}_b$. Given the joint radii and rotations, Eq. \ref{eq:rntilde}-Eq. \ref{eq:rb} provide the FK model of a multi-joint continuum robot that upon solving them recursively, vectors $\vec{t}_i$, $\vec{n}_i$, $\vec{b}_i$ and $\vec{r}_{b,i}$ for all the points on the robot including the end-effector are found. For a three-joint robot, as the one in Fig. \ref{fig:robot_2d}, vector $\vec{r}_{b,4}$ determines the end-effector position. Also, vectors $\vec{t}_4$, $\vec{n}_4$ and $\vec{b}_4$ define the orientation at that location. Note that when $R_i$ tends to infinity, $\phi_i = l_{s,i} / R_{i}$ moves towards zero meaning that Joint $i$ becomes a straight line. In this case, utilizing the aforementioned equations, we conclude

\begin{equation}
    \begin{aligned}
    \vec{t}_i &= \vec{t}_{i-1}, \\
    \vec{n}_i &= \vec{n}_{i-1} \cos(\theta_i) + \vec{b}_{i-1} \sin(\theta_i) \\
    \vec{b}_i &= \vec{b}_{i-1} \cos(\theta_i) - \vec{n}_{i-1} \sin(\theta_i) \\
    \vec{r}_{b,i} &= \vec{r}_{b,i-1} + \vec{t}_{i-1} \left( l_{b,i-1} + l_{s,i-1} + l_{t,i-1}  \right)
    \end{aligned}
\end{equation}


\subsubsection{Inverse Kinematics}

Inverse kinematics (IK) is the inverse of FK where the goal is to find a set of state variables such that the desired end-effector's position and orientation are achieved. In other words
\begin{equation}\label{eq:ik}
\vec{q} = \vec{f}^{-1}_{FK}(\vec{r}_e)
\end{equation}
where $\vec{f}_{FK}$ is the forward kinematic equations presented earlier and $\vec{q}$ denotes the vector of state variables (joints' radii and rotations). Since in this work only the reachability of the robot is considered, vector $\vec{r}_e$ is a three-dimensional vector determining the end-effector's position. Referring to the previous section, for a robot containing $m$ joints, since each joint has two DOF, the robot has $2m$ DOF in total. This means that a robot with only two joints is a redundant manipulator, so Eq. \ref{eq:ik} may have an infinite number of solutions. Many numerical techniques can be applied to solve Eq. \ref{eq:ik} such as the pseudo-inverse, Jacobian transpose or damped least-square method \cite{buss2004introduction}. In this work, the last approach is adopted.

A significant advantage of having redundancy in a robot is that other types of performance criteria (e.g., collision avoidance, actuator torque or energy minimization) can be incorporated in the IK routine. In this paper, minimizing the total actuator torque applied at the base of each continuum joint is selected as the secondary performance measure. Therefore, Eq. \ref{eq:ik} is re-formulated as
\begin{equation}\label{eq:opt}
\begin{aligned}
    \min_{\vec{q}} \quad & \left( \frac{1}{2} \sum_i \left\| \vec{\tau}_i (\vec{q}) \right\|^2 \right) \\
     \mathrm{s.t.} \quad & \quad \: \vec{f}_{FK} (\vec{q}) = \vec{r}_e  \\ & \quad \: \vec{lb} \leq \vec{q} \leq \vec{ub}
\end{aligned}
\end{equation}
where $\vec{\tau_i}$ is the static actuator torque at the base of Joint $i$. Also, $\vec{lb}$ and $\vec{ub}$ specify respectively the desired lower and upper bounds for the joints' radius and rotation. The problem in Eq. \ref{eq:opt} can be solved using a gradient-based constrained optimization method \cite{ebrahimi2019design}. In this work, the constrained trust region algorithm implemented in Python's Scipy package is used. The objective function in Eq. \ref{eq:opt} is the total torque required to hold a given payload at the end-effector location.

\subsection{Optimization Using an EDA}

As stated earlier, an EDA uses a probability model to estimate and sample promising candidate solutions at each iteration. The detailed procedure of the algorithm is as follows.

\subsubsection{Generation of Initial Population} 

An initial population of solutions $\mathbf{P}$ is randomly generated at the start. A solution is a particular instantiation of the design variables $x_i \in \mathcal{D}$ while respecting the dimensional limits per $d_{i, lb} \leq x_i \leq d_{i, ub}$.

\subsubsection{Evaluation and Selection}

Solutions in the population are evaluated by computing the objective and constraint functions. Here, we use a penalty method \cite{smith1997penalty} to quantify the constraint violation and combine it with the objective function to formulate a single fitness function as follows 

\begin{equation}
\begin{aligned}
\text{fitness} := f(\mathcal{D}) + \sigma \max(0, \alpha - \mathbf{\Theta}(\mathcal{D},\mathcal{S}))
\end{aligned}
\end{equation}
where $\sigma$ is a penalty coefficient. Evaluated solutions are ranked based on their fitness values. Then, a subset of the population, $\mathbf{S}$, representing the top $N$ solutions is selected from $\mathbf{P}$. The truncation rate is defined as $|\mathbf{S}| / |\mathbf{P}|$. 

\subsubsection{Estimation of Probability Distribution}

From $\mathbf{S}$, the probability distribution of promising solutions are estimated. For the current work, a univariate normal distribution is used.
\begin{equation}
\begin{aligned}
& p(\mathcal{D}) = \prod_{i=1}^{|\mathcal{D}|} p(x_i), \; x_i \in \mathcal{D} \\
\end{aligned}
\end{equation}
where $p(x_i) \sim \mathcal{N}(\mu_i, \sigma^2_i)$, which assumes that the probability of an optimal design can be independently computed with normally distributed $p(x_i)$. Hence, the estimation of the probability distribution involves computing the mean $\mu_i$ and the variance $\sigma^2_i$ for each variable $x_i$. While more complex probability distributions such as multivariate normal or Gaussian mixture models could be used \cite{larranaga2000optimization}, preliminary experiments showed that a simple model worked better for our case.

\begin{figure*}[!t]
\centering
\subfloat[Mobile platform]{\includegraphics[width=0.3\textwidth]{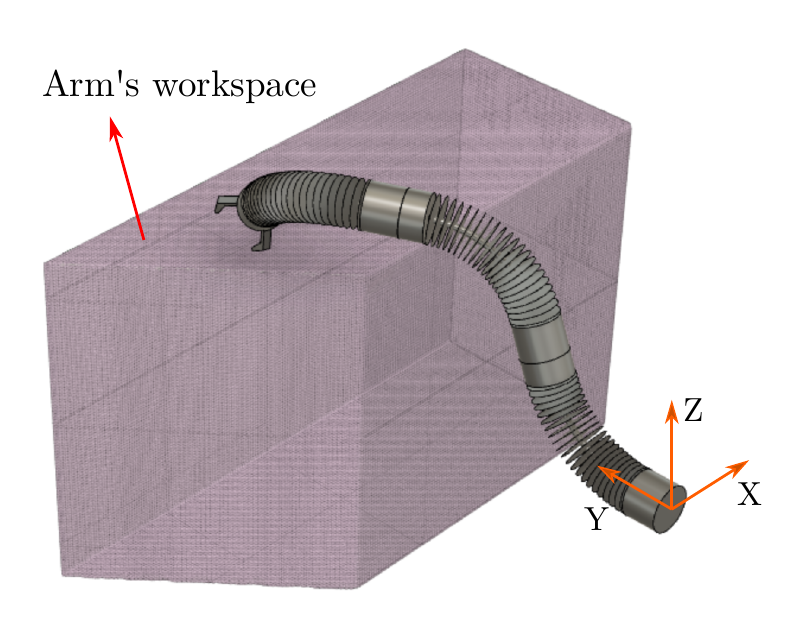}%
\label{fig:problem_1}}
\subfloat[Deep sea mining]{\includegraphics[width=0.3\textwidth]{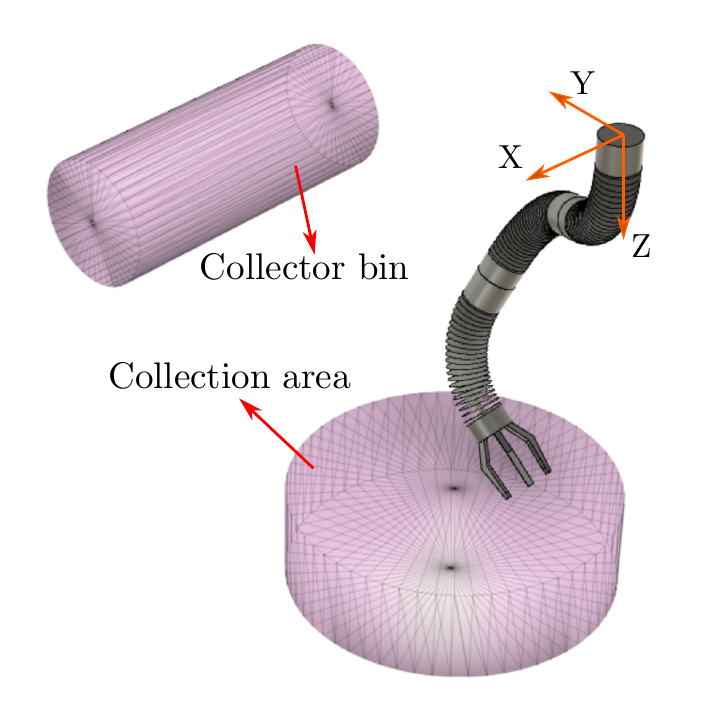}%
\label{fig:problem_2}}
\subfloat[Spot welding]{\includegraphics[width=0.3\textwidth]{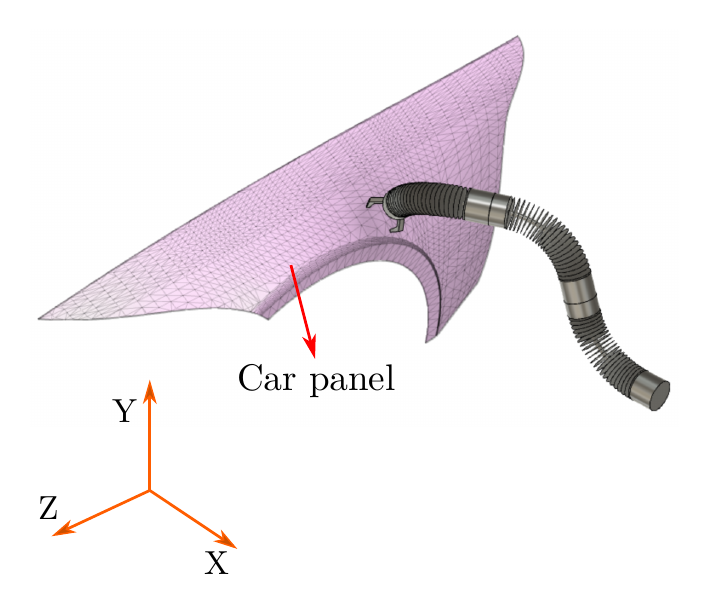}%
\label{fig:problem_3}}
\caption{The robot's location/orientation and the workspaces that must be reached for each application problem. For (a) and (b), the end-effector must reach the entire workspace volumes while for (c), it must reach a series of points along the periphery of the car panel.}
\label{fig:problems}
\end{figure*}

\subsubsection{New Population Generation via Sampling}

A new population of the size $|\mathbf{P}|$ is generated by sampling from the probability distribution estimated from the previous generation of population. This can be performed in a straightforward manner using the Box-Muller transform \cite{box1958note} and the values of $\mu_i$ and $\sigma^2_i$. Note that if necessary, we clip the values of sampled solutions such that the dimensional limits per $d_{i, lb} \leq x_i \leq d_{i, ub}$ in Eq. \ref{eq:opps} are respected.

If the objective function can be computed quickly, such as the total length of the robot, we could repeat sampling until the objective value of each solution in the new population is less than the best objective value found so far. In other words, we only accept a sampled solution for the next generation if
\begin{equation}
\begin{aligned}
f(\mathcal{D}_\text{sampled}) < f_\text{best}
\end{aligned}
\end{equation}
We call this strategy as \textit{select generation} and evaluate its effect on the algorithm's convergence rate in the Experiments section.

\subsubsection{Iterate Steps 2-4}

Steps 2 to 4 are repeated with the newly generated population until meeting a convergence criterion or a maximum number of iterations allocated.

\section{Experiments}

The optimization method is evaluated using three application problems. The first two are problems given to Otherlab Inc. (the last author's affiliate) by its clients while the last problem is created to test the torque minimization capability of the method. The workspaces of each problem can be seen in Fig. \ref{fig:problems}. All dimensional units in this section are in \textit{cm}.

\subsection{Application Problems}

The first application is a two-arm dexterous manipulation on a mobile platform. Two arms are symmetrically mounted on the sides of a vehicle with a camera equidistant away from them along the line of symmetry. The workspace is in the camera's field of view. Fig. \ref{fig:problem_1} represents a portion of the workspace that must be reached by one arm, while the workspace for the second arm (not shown) would be the mirror image of the workspace shown. The envelop dimensions of the workspace including the robot arm's origin are $80 \times 32 \times 51$. The arm's base is positioned at $(0, 0, 0)$ with the tangent direction of the base pointing at $(0, 1, 0)$.

The second application is deep sea mining. The workspace can be seen in Fig. \ref{fig:problem_2}. The robot arm is on a vehicle orientated such that it is pointing towards the ground. The vehicle traverses the ocean floor and the arm picks up nodules and deposits them in a collector. For this application, we represent the workspace as two disjointed volumes; the wide short cylinder being a space above the ocean floor and the thin long cylinder being the collector. The robot only needs to be able to reach each of the volumes but not necessarily the space in between them. The envelop dimensions of the workspace including the both workspace volumes and the arm's base are $51 \times 86 \times 61$. The arm's origin is positioned at $(0, 0, 0)$ with the tangent direction of the base pointing along $(0, 0, 1)$.

The third application is an industrial automation application in which there is a series of points that need to be welded on a car panel. The 3D rendering of the car panel can be seen in Fig. \ref{fig:problem_3}. This application differs from the previous applications because the target workspace is a series of points located on the periphery of the car panel boundary. Thirty-two evenly spaced points are identified along the boundary, with their positions in the ranges of $([13.7, 37.7], [10.9, 70.3], [0.0, -120.8])$. The robot's base is positioned at $(75.0, 45.0, -70.0)$ with its tangent direction at $(-1, 0, 0)$.

For all three application problems, we consider a three-joint continuum robot. The following set of design variables are considered -- the lengths of the first bottom base $l_{b,1}$, first spine $l_{s,1}$, second bottom base $l_{b,2}$, second spine $l_{s,2}$, third bottom base $l_{b,3}$, third spine $l_{s,3}$, and third top base $l_{t,3}$ representing the end-effector. The imposed dimensional constraints on the design variables are
\begin{equation}\label{eq:cons1}
\begin{gathered}
    4 \leq l_{b,1} \leq 30; \;
    2.675 \leq l_{s,1} \leq 32.1 \\
    4 \leq l_{b,2} \leq 30; \;
    2.173 \leq l_{s,2} \leq 26.076 \\
    4 \leq l_{b,3} \leq 30; \;
    2.173 \leq l_{s,3} \leq 26.076; \;
    36 \leq l_{t,3} \leq 60 \\
\end{gathered}
\end{equation}
Note that the top bases for the first and second joints are ignored and assigned zero lengths for optimization, as they are redundant design variables with respect to the bottom bases of the second and third joints. 

In addition, minimum bend radius constraints must be imposed on each joint's radius of curvature ($R_1$, $R_2$ and $R_3$).
\begin{equation}\label{eq:cons2}
\begin{gathered}
    10.22 \leq R_1; \;
    8.3 \leq R_2; \;
    8.3 \leq R_3 \\
\end{gathered}
\end{equation}
A minimum bend radius is the smallest allowed radius that a joint can take while bending, due to its physical limits. These constraints are to be respected during the reachability analysis. For instance, when sampling random configurations with forward kinematics, the values of $R_i$ should be greater than those imposed above. Also, when finding the robot configuration that can reach a particular point using inverse kinematics, the above constraints restrict the possible values of $R_i$ considered.

For the first two problems, the objective function is the total length of the arm, i.e., the sum of all design variables, which is equivalent to having a lightweight robot. The threshold for the reachability constraint is set to 0.95, i.e., the robot must be able to reach 95\% of the target points representing the workspace. For the last problem, the objective function is the total minimum actuator torque required to reach all welding points. The torque is for resisting the gravity load due to the welder device's mass (1 kg) at the end-effector and joints' mass (1 kg each) at their centroids. This torque minimization problem is solved through Eq. \ref{eq:opt} and summing up the minimum actuator torque values $\sum_i \| \vec{\tau}_i (\vec{q}) \|$ for all the welding points. We require that all 32 welding points must be reachable by the robot, i.e., 100\% reachability.

\subsection{Rechability Analysis}

The 3D workspace for each application problem is created using CAD software and exported as an STL model. For the first two application problems, we convert their STL models into 3D voxels (capturing the interior volumes) with each voxel sized at 3\textit{cm} $\times$ 3\textit{cm} $\times$ 3\textit{cm}. Using forward kinematics, if the end-effector's position is within a particular voxel, we conclude that the target point represented by that voxel can be reached. As for inverse kinematics, we solve for possible robot configurations that would result in the end-effector's position coinciding with the center of each target voxel with the tolerance $\epsilon$ of 1\textit{cm}. The lower tolerance is used for the inverse kinematics because more accurate evaluation of reachability is desired for solutions that are close to the feasibility threshold.

\begin{figure}[!tb]
\centering
\subfloat[]{\includegraphics[width=0.23\textwidth]{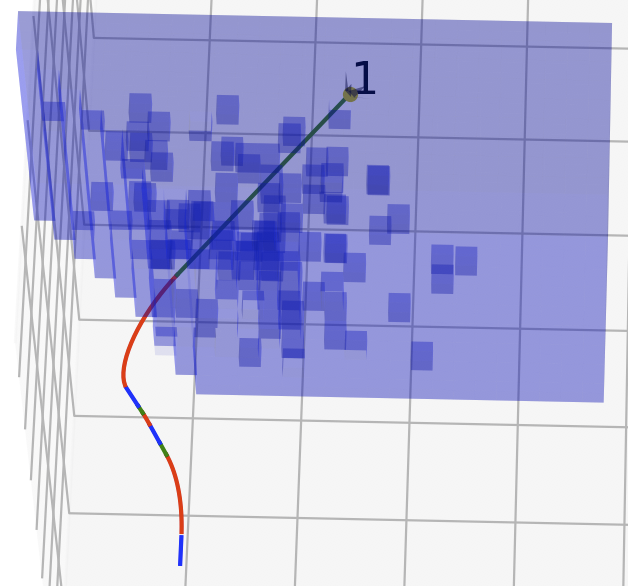}%
\label{fig:reach_1}}
\subfloat[]{\includegraphics[width=0.23\textwidth]{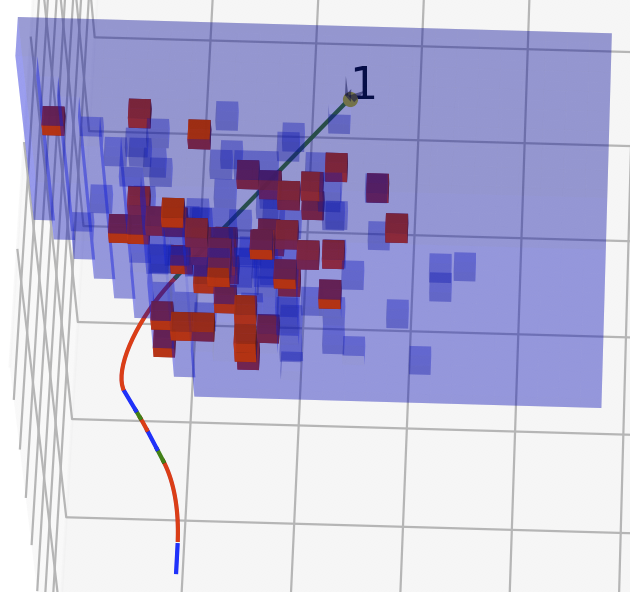}%
\label{fig:reach_2}}
\caption{Reachability analysis performed using the hybrid approach. (a) First, the reachability is estimated using forward kinematics with randomly generated configurations. (b) Then, inverse kinematics is used to check the reachability of those points not reached during the forward kinematics simulation. Dark blue boxes indicate the points deemed unreachable within the workspace by the forward kinematics simulation. Red boxes indicate the missed points confirmed as reachable by the inverse kinematics simulation.}
\label{fig:reach}
\end{figure}

For the first two application problems, we use the hybrid approach of incorporating both forward kinematics and inverse kinematics to estimate the reachability (illustrated in Fig. \ref{fig:reach}). First, three million robot configurations (as recommend in \cite{bodily2017multi}) are generated by randomly sampling the values of $R_i$ and $\theta_i$, and forward kinematics for each configuration is performed to estimate the reachability of a given candidate solution. Then, inverse kinematics is used if the estimated reachability by the forward kinematics approach does not meet but is close to the required threshold. Here, we utilize inverse kinematics to check on the points that are not reached during the forward kinematics simulation if the estimated reachability is between 0.9 and 0.95 (the required threshold). This hybrid approach prevents the algorithm from unnecessarily penalizing a feasible solution, the reachability of which could be underestimated if only the forward kinematics approach is used.

For the last application problem, the 32 welding points are identified along the boundary of the 3D car panel model. We use inverse kinematics to check whether all welding points can be reached or not, while simultaneously solving for the minimum torque required to reach all those points.

\begin{figure*}[!t]
\centering
\subfloat[Mobile platform]{\includegraphics[width=0.333\textwidth]{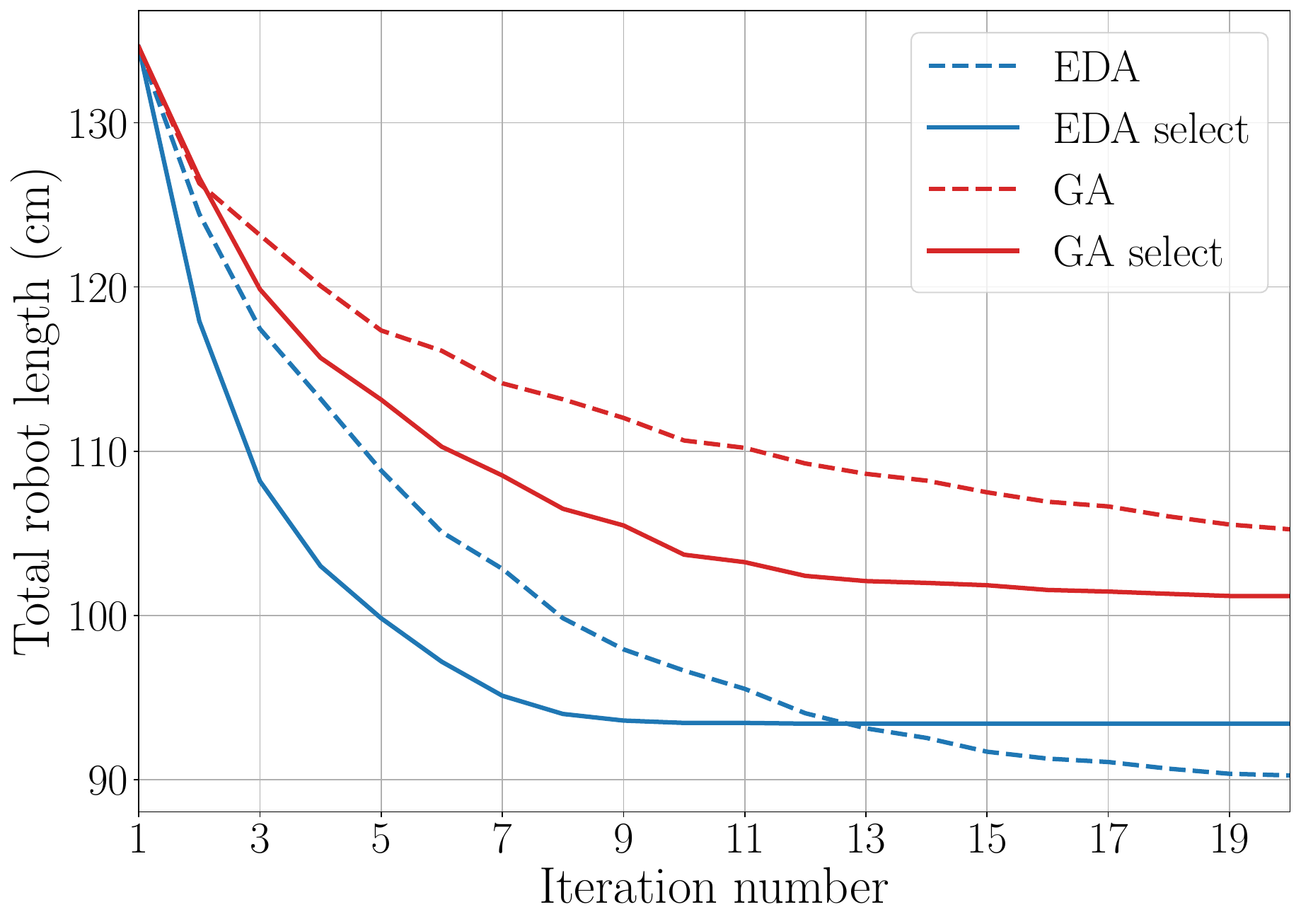}%
\label{fig:results1}}
\subfloat[Deep sea mining]{\includegraphics[width=0.333\textwidth]{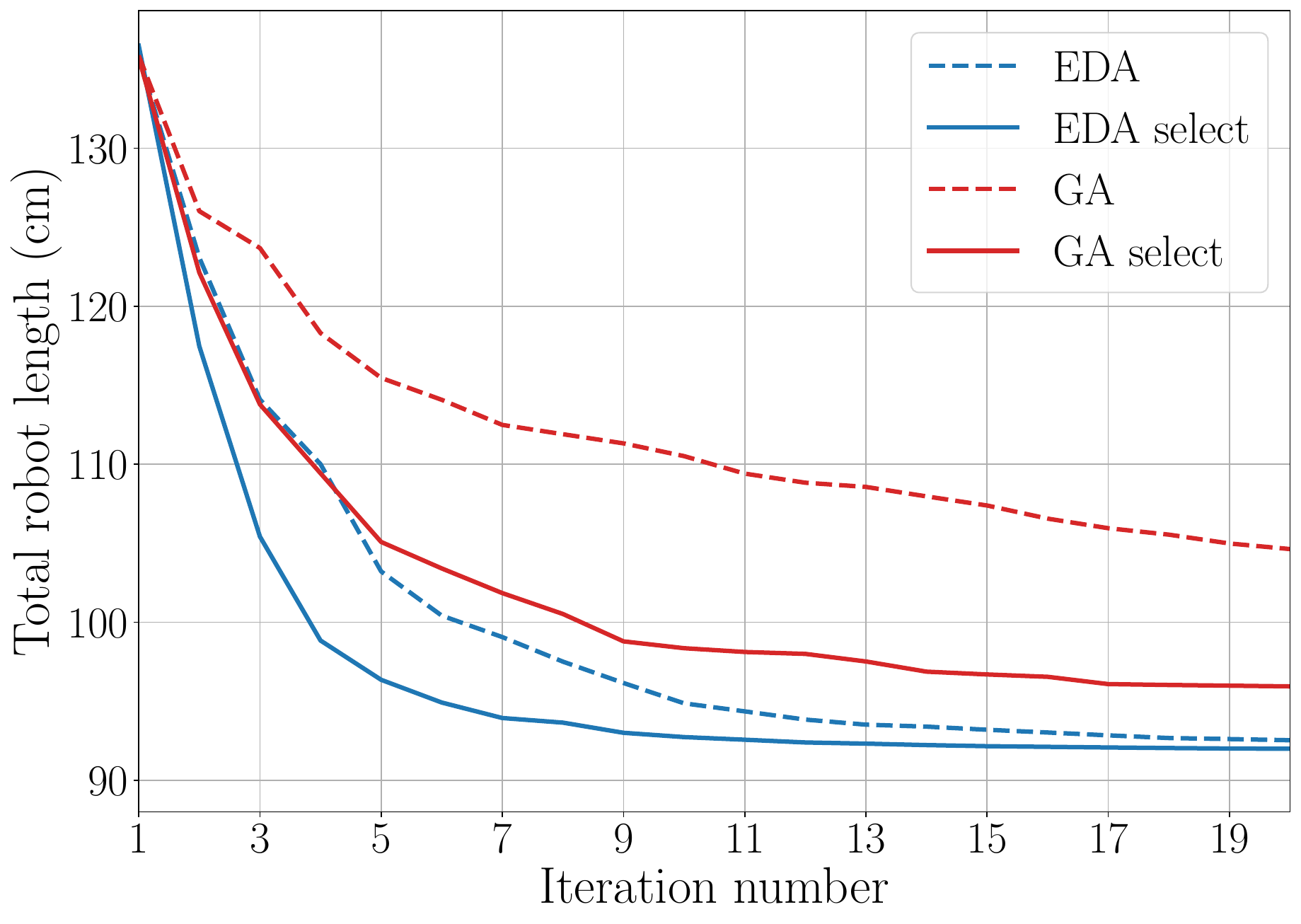}%
\label{fig:results2}}
\subfloat[Spot welding]{\includegraphics[width=0.333\textwidth]{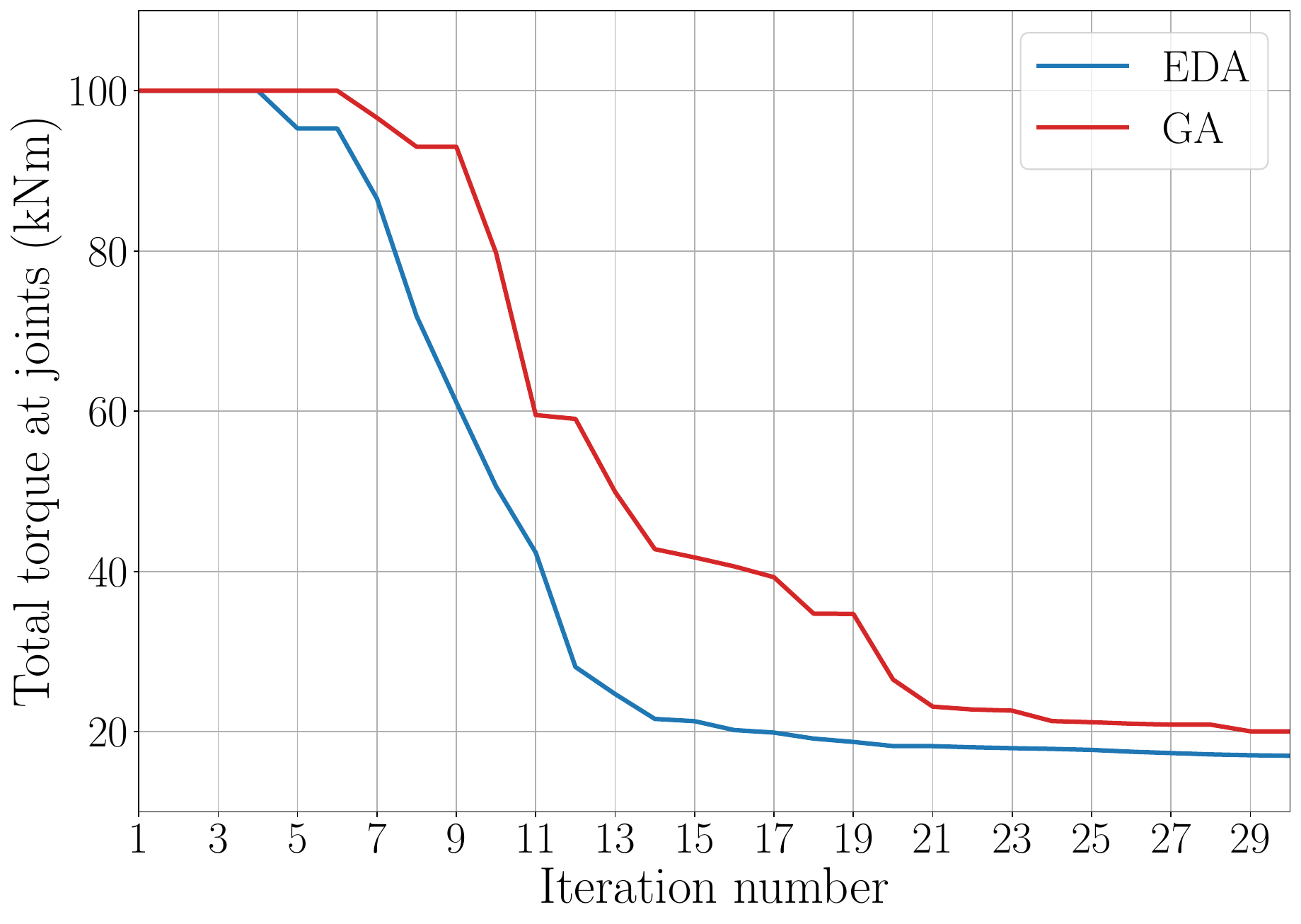}%
\label{fig:results3}}
\caption{Summary of optimization results for the application problems. Reported in the y-axis are the objective values of the best feasible solution at each iteration, averaged from 20 runs for each algorithm. The algorithms with "select" notation indicate the usage of select generation. In all cases, the final objective values are significantly lower for the EDA compared to the GA.}
\label{fig:results}
\end{figure*}

\subsection{Algorithms Compared}

We compare the EDA based on a univariate normal distribution against a genetic algorithm (GA) for all three application problems. The GA is considered as the benchmark method because it was used in prior work to optimize a multi-joint continuum robot considering reachability \cite{bodily2017multi}. The cross-over and mutation operators used for the GA are the same as those in \cite{bodily2017multi}, with the cross-over and mutation rates of 0.9 and 0.1, respectively. For selection, both algorithms incorporate the same truncation method with the truncation rate of 0.5. 

For the first two application problems, we test two versions of each EDA/GA. The first version of each algorithm uses the standard generation procedure at each iteration. On the other hand, the second version of each algorithm incorporates the select generation strategy described in the Methods section. The strategy involves accepting new solutions -- sampled from probability distributions for the EDA vs. created using genetic operators for the GA -- only if their objective value is smaller than the best objective value found hitherto. Select generation is repeated until the required population size is met or the maximum number of trials ($=$10000) is reached, at which point the standard generation procedure is used to fill the remaining population.

For the last application problem, the evaluation of the objective function requires computing inverse kinematics for all target points, which is much more expensive than computing the total length as in the first two problems. Hence, we do not examine the select generation strategy and compare the EDA against the GA only with the standard generation procedure.

For all algorithms, the population size of 100 is used. The maximum iteration number allocated is 20 for the first two problems and 30 for the last one. The penalty coefficient used for the reachability constraint is 0.33 for all three problems. Each algorithm is repeated 20 times, starting with the same randomly generated initial population for each run. For the results, the average objective values of the best feasible solution (i.e., the solution that meets the reachability constraint with the lowest objective value) at each iteration are reported. 

\subsection{Results}

Fig. \ref{fig:results} summarizes the optimization results and Table \ref{tab1} reports the mean objective values and their standard deviations found at the end of optimization runs. For all three application problems, the EDA outperforms the GA in terms of the quality of the best solutions found at the end. The final objective values found are significantly lower with the EDA than the GA ($p<0.0001$ for all cases, including with or without select generation). Note that the standard deviations of the final objective values found by the EDAs are very low, indicating a convergence to optimal solutions at the end.

In general, the select generation strategy seems to improve the convergence rate for both the EDA and the GA, as shown in Fig. \ref{fig:results}. However, for the first application problem, it causes the EDA to get stuck in a local minimum and the version with standard generation finds better solutions in the end ($p<0.0001$) than the version with select generation. For the second application problem, the select generation strategy resulted in significantly lower final objective values for the GA compared to the standard generation strategy ($p<0.0001$). 

\begin{table}[htbp]
\caption{Final objective values found for each setting}
\begin{center}
\begin{tabular}{|c|c|c|c|c|c|c|}
\hline
\multirow{2}{*}{} & \multicolumn{2}{c|}{\textbf{Mobile platform}} & \multicolumn{2}{c|}{\textbf{Deep sea mining}} & \multicolumn{2}{c|}{\textbf{Spot welding}} \\ \cline{2-7} 
 & \textit{\textbf{\begin{tabular}[c]{@{}c@{}}mean \\ (cm)\end{tabular}}} & \textit{\textbf{SD}} & \textit{\textbf{\begin{tabular}[c]{@{}c@{}}mean\\ (cm)\end{tabular}}} & \textit{\textbf{SD}} & \textit{\textbf{\begin{tabular}[c]{@{}c@{}}mean\\ (kNm)\end{tabular}}} & \textit{\textbf{SD}} \\ \hline
\textbf{EDA} & 90.3 & 0.64 & 92.5 & 0.69 & 17.0 & 1.8 \\ \hline
\textbf{GA} & 105.3 & 3.6 & 104.6 & 3.0 & 20.1 & 2.5 \\ \hline
\textbf{EDA-select} & 93.4 & 3.0 & 92 & 0.20 &  &  \\ \hline
\textbf{GA-select} & 101.2 & 5.2 & 95.9 & 2.6 &  &  \\ \hline
\end{tabular}
\label{tab1}
\end{center}
\end{table}

In terms of computation time, each optimization run takes about 3.5 hours for the first two problems and 1.7 hours for the last problem, on a single desktop computer with two Intel Xeon CPUs (E5-2650 v2 2.60 GHz) and 32GB of RAM. Note that because the EDA/GA is a population-based algorithm, we could run 16 evaluations in parallel during optimization runs. The computation time for both the EDA and the GA are roughly the same because the majority of the time is spent on evaluating solutions vs. other procedures in the algorithms.

In summary, the overall results of our experiments demonstrate the effectiveness of the EDA compared to the GA in finding optimal designs of continuum robots considering reachability constraints. In addition, the select sampling strategy has shown the potential in improving the convergence rate, although it could lead to premature convergence in some cases.

\section{Summary and Conclusions}

The current work has presented a computational method to find optimal designs of continuum robots while considering reachability constraints. To assess the reachability of a given robot design, the method takes advantage of both forward kinematic and inverse kinematic approaches. The former with randomly sampled robot configurations is used to quickly estimate the reachability, while the latter is used to further assess the reachability so that a feasible solution could be more accurately identified during optimization. 

In addition, our implementation of inverse kinematics can incorporate the minimization of secondary performance criteria, specifically in the current work the minimum torque required to reach a target point. This capability allows us to find an optimal design with the total minimum actuator torque required to reach the workspace. 

Lastly, the optimization method is based on the estimation of distribution algorithm (EDA), a population-based, derivative-free optimization method that uses a univariate marginal distribution to estimate and sample promising candidate solutions. It also features a penalty method to handle reachability constraints and a select generation strategy to increase the convergence rate. Through three application problems, the EDA is shown to be superior to the genetic algorithm implemented in finding better solutions within a given number of iterations. In practice, the method could find optimal solutions in 2-4 hours rather than the typical 1-2 weeks taken for the manual work performed by an engineering team. This would drastically decrease the overall design time and also allow the team to consider a greater number of alternatives for the final design.

Future work includes considering other performance criteria of continuum robots, such as their dexterity and manipulability. The scalability of the proposed method should be investigated as well, e.g., optimizing robots with more joints. Also, other derivative-free optimization algorithms or reinforcement learning techniques can be explored for solving the problem. Lastly, the current method using the EDA could be extended to solve other optimal design problems found in robotics.

\bibliographystyle{IEEEtran}
\bibliography{IEEEabrv,bibfile}

\end{document}